%% file: main.tex
\documentclass[conference]{IEEEtran}
\IEEEoverridecommandlockouts
\usepackage{cite}
\usepackage{amsmath,amssymb,amsfonts}
\usepackage{algorithmic}
\usepackage{graphicx}
\usepackage{textcomp}
\usepackage{xcolor}
\usepackage[caption=false,font=normalsize,labelfont=sf,textfont=sf]{subfig}
\usepackage{booktabs,threeparttable}
\def\BibTeX{{\rm B\kern-.05em{\sc i\kern-.025em b}\kern-.08em
    T\kern-.1667em\lower.7ex\hbox{E}\kern-.125emX}}

\allowdisplaybreaks
\usepackage{balance}
\usepackage{adjustbox}
\usepackage[thinlines]{easytable}
\usepackage{gensymb}

\newcommand{\Renyi}{R\'{e}nyi }
\newcommand\subscr[2]{#1_{\textup{#2}}}

\newcommand{\accgap}{\subscr{{acc}}{gap}}
\newcommand{\accmin}{\subscr{{acc}}{min}}
\newcommand{\dpgap}{\subscr{dp}{gap}}
\newcommand{\eqoddsgap}{\subscr{{eqodds}}{gap}}
    
\makeatletter
\newcommand{\newlineauthors}{%
  \end{@IEEEauthorhalign}\hfill\mbox{}\par
  \mbox{}\hfill\begin{@IEEEauthorhalign}
}
\makeatother

\begin{document}

\title{\Renyi Fair Information Bottleneck \\for Image Classification
\thanks{This work was supported in part by NSERC of Canada.}
}

\author{\IEEEauthorblockN{Adam Gronowski}
\IEEEauthorblockA{\textit{Dept.\ Mathematics and Statistics} \\
\textit{Queen's University}\\
Kingston, ON, Canada \\
adam.gronowski@queensu.ca}
\and
\IEEEauthorblockN{William Paul}
\IEEEauthorblockA{\textit{Applied Physics Laboratory} \\
\textit{Johns Hopkins University}\\
Laurel, MD, United States \\
william.paul@jhuapl.edu}
\and
\IEEEauthorblockN{Fady Alajaji}
\IEEEauthorblockA{\textit{Dept.\ Mathematics and Statistics} \\
\textit{Queen's University}\\
Kingston, ON, Canada \\
fa@queensu.ca}
\newlineauthors
\IEEEauthorblockN{Bahman Gharesifard}
\IEEEauthorblockA{\textit{Electrical \& Computer Engineering Dept.} \\
\textit{University of California, Los Angeles}\\
Los Angeles, CA, United States \\
gharesifard@ucla.edu}
\and
\IEEEauthorblockN{Philippe Burlina}
\IEEEauthorblockA{\textit{Applied Physics Laboratory} \\
\textit{Johns Hopkins University}\\
Laurel, MD, United States \\
philippe.burlina@jhuapl.edu}
}

\maketitle

\begin{abstract}
We develop a novel method for ensuring fairness in machine learning which we term as the \Renyi Fair Information Bottleneck (RFIB). We consider two different fairness constraints - demographic parity and equalized odds - for learning fair representations and derive a loss function via a variational approach that uses R\'enyi's divergence with its tunable parameter \boldmath{$\alpha$} and that takes into account the triple constraints of utility, fairness, and compactness of representation. We then evaluate the performance of our method for image classification using the EyePACS medical imaging dataset, showing it outperforms competing state of the art techniques with performance  measured  using  a  variety  of  compound  utility/fairness metrics, including accuracy gap and Rawls’ minimal accuracy.

\end{abstract}


\input{introduction}

\input{methods}
\input{experiments}

\input{discussion}

\balance
{\small
\bibliographystyle{IEEEtran}
\bibliography{egbib}
}

\end{document}

%% file: introduction.tex
\section{Introduction}

The problem of fairness in machine learning is to obtain accurate predictions of a target of interest while remaining free of bias due to sensitive information such as gender, race, age, or other similar attributes. Representing input data as random variable $X \in \mathcal{X}$,  prediction target as random variable $Y \in \mathcal{Y}$, and sensitive information as random variable $S \in \mathcal{S}$, the goal is to predict $Y$ from $X$ in a way that is uninfluenced by $S$. 

One solution to this problem is to learn \textit{fair representations}, finding an intermediate representation $Z \in \mathcal{Z}$ that can then be used instead of $X$ with existing machine learning architectures to make predictions, as done by \cite{zemel2013learning,song2019learning}. The new representation $Z$ must simultaneously preserve information from $X$ relevant to predicting $Y$ while removing sensitive information that could lead to bias.

Learning fair representations can be done both by adversarial methods, such as \cite{beutel2017data,edwards2015censoring,madras2018learning}, and through variational approaches such as \cite{skoglund2020,ghassami2018fairness} that  maximize and minimize different mutual information terms to ensure the representation is both expressive and fair. These approaches are related to the \textit{information bottleneck (IB)} method \cite{tishby1999information} and its variational approximations \cite{deepvariational,fischer2020conditional,kolchinsky2019nonlinear} which aim to find a representation that maximally compresses $X$ while remaining informative about $Y.$ 

We develop a related variational IB method for fairness using \Renyi divergence due to the extra degree of freedom provided by its tunable $\alpha$ parameter, rather than the more commonly used Kullback-Leibler divergence. We learn representations that fulfill three criteria: the representation is compact, expressive about $Y$ in order to improve classification accuracy, and free from information that can lead to bias. Unlike most prior work that focuses on a single definition of fairness, we consider how to jointly address and balance two of arguably the most important measures for
fairness, demographic parity and equalized odds, with possible benefits  for  ethicists  and  policy makers  to  translate  policies into actual  engineering  systems. 

Recent work based on \Renyi information measures (and its variants) include an IB problem under a \Renyi entropy complexity constraint \cite{weng2021information}, bounding the generalization error in learning algorithms~\cite{esposito20}, \Renyi divergence variational inference~\cite{varInf}, \Renyi differential privacy \cite{mironov2017renyi} and the analysis and development of deep generative adversarial networks \cite{bhatia2021least, rgan, cumulant_gan, kurri2021realizing}.

The rest of the paper is organized as follows. In Section II we derive a cost function for our method to be minimized and then in Section III present experimental results using our method to classify retinal fundus images from the EyePACS dataset. Finally, we conclude with a discussion in Section IV.

%% file: methods.tex
\section{R\'enyi Fair Information Bottleneck}
We adopt a variational approach which we call {\em R\'enyi Fair Information Bottleneck} (RFIB) to encode data $X$ into a new representation $Z$ that can be used to draw inferences about $Y$ free from bias due to $S$. In light of our model, we assume that the Markov chain $(Y,S) \rightarrow X \rightarrow Z$ holds. To simplify notation, we assume in this section that all random variables are discrete, though a similar derivation holds for a mix of continuous and discrete random variables.

\subsection{Fairness Defined}
Among the three principal definitions of fairness -- \emph{demographic parity}, \emph{equalized odds}, and \emph{equality of opportunity} -- we focus on addressing both demographic parity and equalized odds since a) equalized odds is related to, but a stronger constraint than, equality of opportunity, and b) demographic parity, also called statistical parity, is an altogether different type of constraint compared to the former two constraints in that the requirement of independence does not involve the actual target label value.

For demographic parity, the goal is for the model's prediction $\hat{Y}$ to be independent of the sensitive variable $S$, i.e.,
\begin{equation}
P(\hat{Y}=\hat{y})=P(\hat{Y}=\hat{y} \mid S=s), \, \forall s,\hat{y},
\end{equation}
while for equalized odds the goal is to achieve this independence by conditioning on the actual target $Y,$ i.e.,
\begin{equation}
P(\hat{Y}=\hat{y}\mid Y=y)=P(\hat{Y}=\hat{y}\mid S=s,Y=y),\forall s,\hat{y},y.
\end{equation}

\subsection{Lagrangian Formulation}
To encourage equalized odds, we minimize $I(Z;S|Y)$; i.e., we minimize the average amount of information that $Z$ has about $S$ given $Y.$ To both obtain good classification accuracy and help promote demographic parity, we maximize $I(Z;Y|S)$. Maximizing mutual information between $Z$ and $Y$ ensures the representation will be expressive about its target while the conditioning on $S$ ensures that $Z$ does not keep information shared by $S$, encouraging demographic parity.


 In addition, we minimize $I(Z;X|S,Y)$, a compression term similar to one from the IB problem~\cite{deepvariational}. This minimization further encourages $Z$ to discard information irrelevant for drawing predictions about $Y,$ hence improving generalization capability and reducing the risk of keeping nuisances.
Finally, we maximize the utility term $I(Z;Y)$; this optimization, similar to the IB problem, solely ensures the representation is maximally expressive of the target $Y.$

Combining these terms leads to a Lagrangian, ${\cal L}$, that we seek to minimize over the encoding conditional distribution $P_{Z|X}$. The Lagrangian is given by
\begin{eqnarray}
    \cal{L} & = & I(Z;S|Y) + I(Z;X|S,Y) -\lambda_1 I(Z;Y) \nonumber \\
    & &  -\lambda_2 I(Z;Y|S),
\end{eqnarray}
where $\lambda_1$ and $\lambda_2$ are hyperparameters.
Reworking this Lagrangian, we have that:
\begin{align}
\cal{L} &= H(Z|Y) - H(Z|S,Y) +H(Z|S,Y) \nonumber \\&\quad- H(Z|X,S,Y) - \lambda_1I(Z;Y) - \lambda_2 I(Z;Y|S)  \nonumber \\
& = H(Z|Y) - H(Z|X) -  \lambda_1I(Z;Y) - \lambda_2 I(Z;Y|S) \nonumber \\
\begin{split}
&= H(X) - H(Z,X) - [H(Y) - H(Z,Y)]\\&\quad -\lambda_1I(Z;Y) - \lambda_2 I(Z;Y|S) \nonumber \\
\end{split}
\\
&= I(Z;X) - I(Z;Y) - \lambda_1I(Z;Y) - \lambda_2 I(Z;Y|S) \nonumber \\
&= I(Z;X) - (\lambda_1 + 1)I(Z;Y) - \lambda_2 I(Z;Y|S), 
\end{align}
where $H(\cdot)$ denotes entropy, and the second equality follows from the Markov chain assumption $(Y,S) \rightarrow X \rightarrow Z$. Hence, we have shown that the Langrangian $\mathcal{L}$ admits a simpler equivalent expression given by
\begin{equation}\label{simple-L}
\mathcal{L} = I(Z;X) -\beta_1 I(Z;Y) -\beta_2 I(Z;Y|S),
\end{equation}
where $\beta_1 = \lambda_1 + 1$ and $\beta_2 = \lambda_1$.
This simpler Lagrangian is easier to compute while maintaining similar properties to the original one. It also reveals a direct relation of the original Lagrangian with the first two terms being exactly equivalent to the ``classical IB'' formulation. The two hyperparameters $\beta_1$ and $\beta_2$ control trade-offs between accuracy and fairness, with higher $\beta$ values corresponding to a higher priority on accuracy and lower $\beta$ values giving more influence to the compression term $I(Z;X)$ that discards unwanted information, potentially improving fairness at the expense of accuracy. As $I(Z;Y)$ is partially derived from the $I(Z;S|Y)$ term designed to improve equalized odds, using a higher $\beta_1$ over $\beta_2$ should give more priority to improving equalized odds, whereas a higher $\beta_2$ should result in improved demographic parity. This allows for more nuanced outcomes compared to other methods that focus rigidly on a single fairness metric. It is also possibly an interesting tool for policy makers to translate those more balanced and nuanced versions of fairness into an ``engineered system.''

\subsection{Variational Bounds} We use a variational approach to develop bounds on the three terms in the Lagrangian in \eqref{simple-L}, finding lower bounds for the terms to be maximized and an upper bound for the term to be minimized. The Markov chain property $(Y,S) \rightarrow X \rightarrow Z$ results in the joint distribution $P_{SYXZ}$ factoring as as $P_{SYX}P_{Z|X}.$ 

The distribution $P_{Z|X}$ is a parametric stochastic encoder to be designed while all other distributions are fully determined by the joint data distribution $P_{S,X,Y}$, the encoder, and the Markov chain constraint. To simplify notation, we simply write $P_{Z|X}$ rather than including the parameter $P_{Z|X,\theta}$, with $\theta$ denoting network weights. Computing the mutual information terms requires the usually intractable distributions $P_{Y|S,Z}$, $P_{Y|Z}$, and $P_{Z}$ so we replace them with variational approximations $Q_{Y|S,Z}$, $Q_{Y|Z}$ and $Q_{Z}$, respectively. We next derive an upper bound for $I(Z;X)$ with the novel use of R\'enyi divergence: 
\begin{align}\nonumber
 I(Z;X) &= \sum_{(z,x) \in \mathcal{Z} \times \mathcal{X}} \hspace{-0.15in} P_{Z,X}(z,x)\log \frac{P_{Z|X}(z|x)}{P_Z(z)} \nonumber\\  
\begin{split}
&= \sum_{(z,x) \in \mathcal{Z} \times \mathcal{X}} \hspace{-0.15in} P_{Z,X}(z,x)\log P_{Z|X}(z|x) \\ &\quad- D_{KL}(P_Z||Q_Z) - \sum_{z\in \mathcal{Z}} P_Z(z)\log Q_Z(z) \\
\end{split}\nonumber
\\ 
&\leq \sum_{(z,x) \in \mathcal{Z} \times \mathcal{X}} 
\hspace{-0.15in} P_{Z,X}(z,x)\log \frac{P_{Z|X}(z|x)}{Q_Z(z)} 
\nonumber\\  
& = \mathbb{E}_{P_X} D_{KL}\left(P_{Z \mid X} \| Q_{Z}\right)\nonumber \\ 
& \leq \mathbb{E}_{P_X} D_\alpha\left(P_{Z \mid X} \| Q_{Z}\right), \label{renyi-bd}
\end{align}
for $\alpha >1$.
The first inequality follows from the non-negativity of Kullback-Leibler (KL) divergence, similar to~\cite{deepvariational,song2019learning,skoglund2020}. For the final step, we take the \Renyi divergence $D_\alpha(\cdot \| \cdot)$ of order $\alpha$ (e.g., see~\cite{renyiDivergence}), rather than the KL divergence as typically done in the literature, where
\begin{equation}
    D_\alpha(P||Q) = \frac{1}{\alpha-1} \log \left(\sum_{x\in \mathcal{X}} P(x)^{\alpha}Q(x)^{1-\alpha}  \right)
\end{equation}
for $\alpha>0$, $\alpha \ne 1$ and distributions $P$ and $Q$ with common support~$\mathcal{X}$.\footnote{If $P$ and $Q$ are probability density functions, then $D_\alpha(P||Q) = \frac{1}{\alpha-1} \log \left( \int_{\mathcal{X}} P(x)^{\alpha}Q(x)^{1-\alpha} \, dx \right)$.}
Using \Renyi divergence gives an extra degree of freedom and allows more control over the compression term~$I(X;Z)$. As the R\'enyi divergence is non-decreasing with $\alpha$, a higher $\alpha$ will more strongly force the distribution $P_{Z|X}$ closer to $Q_Z$, resulting in more compression. 

The upper bound in \eqref{renyi-bd} holds for $\alpha > 1$ since $D_\alpha$ is non-decreasing in $\alpha$ and
$\lim_{\alpha \to 1} D_\alpha(P\|Q) = D_{KL}(P\|Q)$.\footnote{For simplicity and by the continuity property of $D_\alpha$, we define its extended orders at $\alpha=1$ and $\alpha=0$ \cite{renyiDivergence} as
$D_1(P\|Q) := D_{KL}(P\|Q)$ and
$D_0(P\|Q) :=\lim_{\alpha \to 0} D_\alpha(P\|Q) = -\log Q(x: P(x)>0)$, which is equal to 0 when $P$ and $Q$ share a common support.} When $\alpha < 1$, then $\mathbb{E}_{P_X} D_\alpha\big(P_{Z \mid X} \| Q_{Z}\big)$ is no longer an upper bound on $I(Z;X)$; but it can be considered as a potentially useful approximation that is tunable by varying~$\alpha$.

We can similarly leverage the non-negativity of KL divergence to get lower bounds on $I(Z;Y)$ and $I(Z;Y|S)$:
\begin{equation}
I(Z;Y) \geq \mathbb{E}_{P_{Y,Z}} \left[ \log Q_{Y | Z}(Y | Z)\right]+H(Y),
\end{equation}
\begin{equation}
\hspace{-0.09in} I(Z;Y|S) \geq \mathbb{E}_{P_{S,Y,Z}} \hspace{-0.03in} \big[\log Q_{Y | S,Z }(Y |S,Z)\big]+H(Y|S).
\end{equation}
As the entropy $H(Y)$ and conditional entropy $H(Y|S)$ of the labels do not depend on the parameterization they can be ignored for the optimization. 

\subsection{Computing the Bounds}
To compute the bounds in practice we use the reparameterization trick \cite{kingma2013auto}. Modeling $P_{Z|X}$ as a density, we let $P_{Z|X}dZ$ = $P_E dE$ where $E$ is a random variable and $Z=f(X,E)$ is a deterministic function, allowing us to backpropagate gradients and optimize the parameter via gradient descent. We use the data's empirical densities to estimate  
$P_{X,S}$ and $P_{X,Y,S}$.

Considering a batch $D = \{x_i,s_i,y_i \}_{i=1}^N$ this finally leads to the following RFIB cost function to minimize:
\begin{equation}\label{final-loss}
\begin{split}
    J_\text{RFIB} &= \frac{1}{N}\sum_{i=1}^{N} \Big[ D_{\alpha}(P_{Z|X=x_i}||Q_Z) \\&\quad- \beta_1 \mathbb{E}_E\left[\log \left(Q_{Y|Z}\left(y_i|f(x_i,E)\right)  \right)  \right]\\ &\quad- \beta_2 \mathbb{E}_E\left[\log \left(Q_{Y|S,Z}\left(y_i|s_i,f(x_i,E)\right)  \right)  \right]\Big],
\end{split}
\end{equation}
where we estimate the expectation over $E$ using a single Monte Carlo sample.

We note that depending on the choice of $\alpha$, $\beta_1$, and $\beta_2$, from our method we can recover both the \emph{IB} \cite{deepvariational} and \emph{conditional fairness bottleneck} (CFB) \cite{skoglund2020} schemes to which we compare our results. Letting $\alpha = 1$ and $\beta_2 = 0$ corresponds to IB, while setting $\alpha = 1$ and $\beta_1 = 0$ corresponds to CFB.

%% file: experiments.tex
\section{Experiments}
We present experimental results on the EyePACS dataset of retinal images.

\subsection{Data}
The EyePACS dataset \cite{eyepacs} is sourced from the Kaggle Diabetic Retinopathy challenge. It consists of 88,692 retinal fundus images of individuals potentially suffering from diabetic retinopathy (DR), an eye disease associated with diabetes that is one of the leading causes of visual impairment worldwide. The dataset contains 5 categories of images based on the severity of the disease, with 0 being completely healthy and 4 being the most severe form of the disease. Similar to \cite{paul2020tara}, we binarize this label into our prediction target $Y,$ with $Y = 1$ corresponding to categories 1-4, considered a positive, referable case for DR, and $Y = 0$ corresponding to category 0, a healthy eye with no disease.

\begin{figure}[hbt]
\centering
\centering
\subfloat{\includegraphics[width=.17\textwidth]{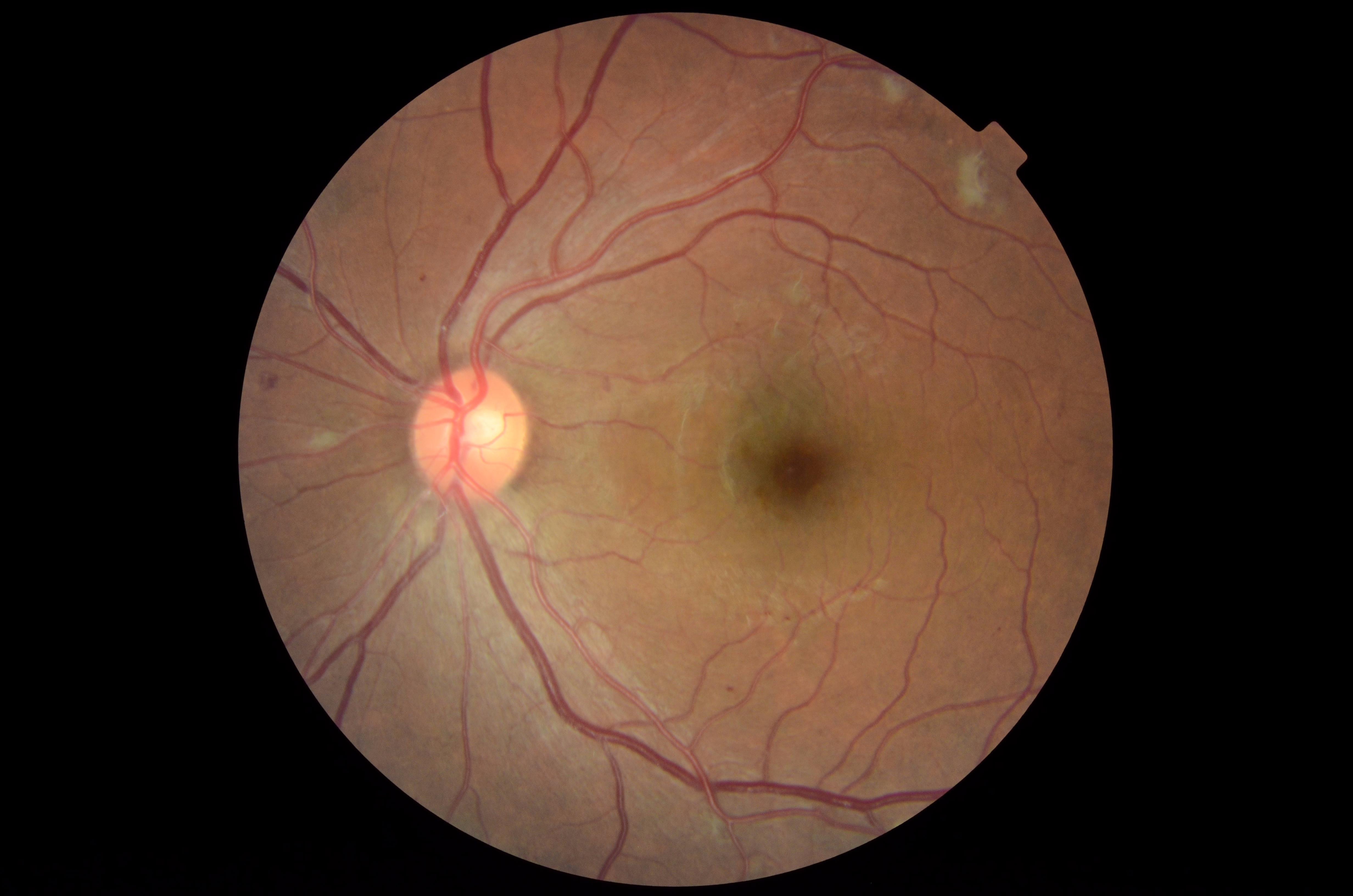}%
}
\subfloat{\includegraphics[width=.15\textwidth]{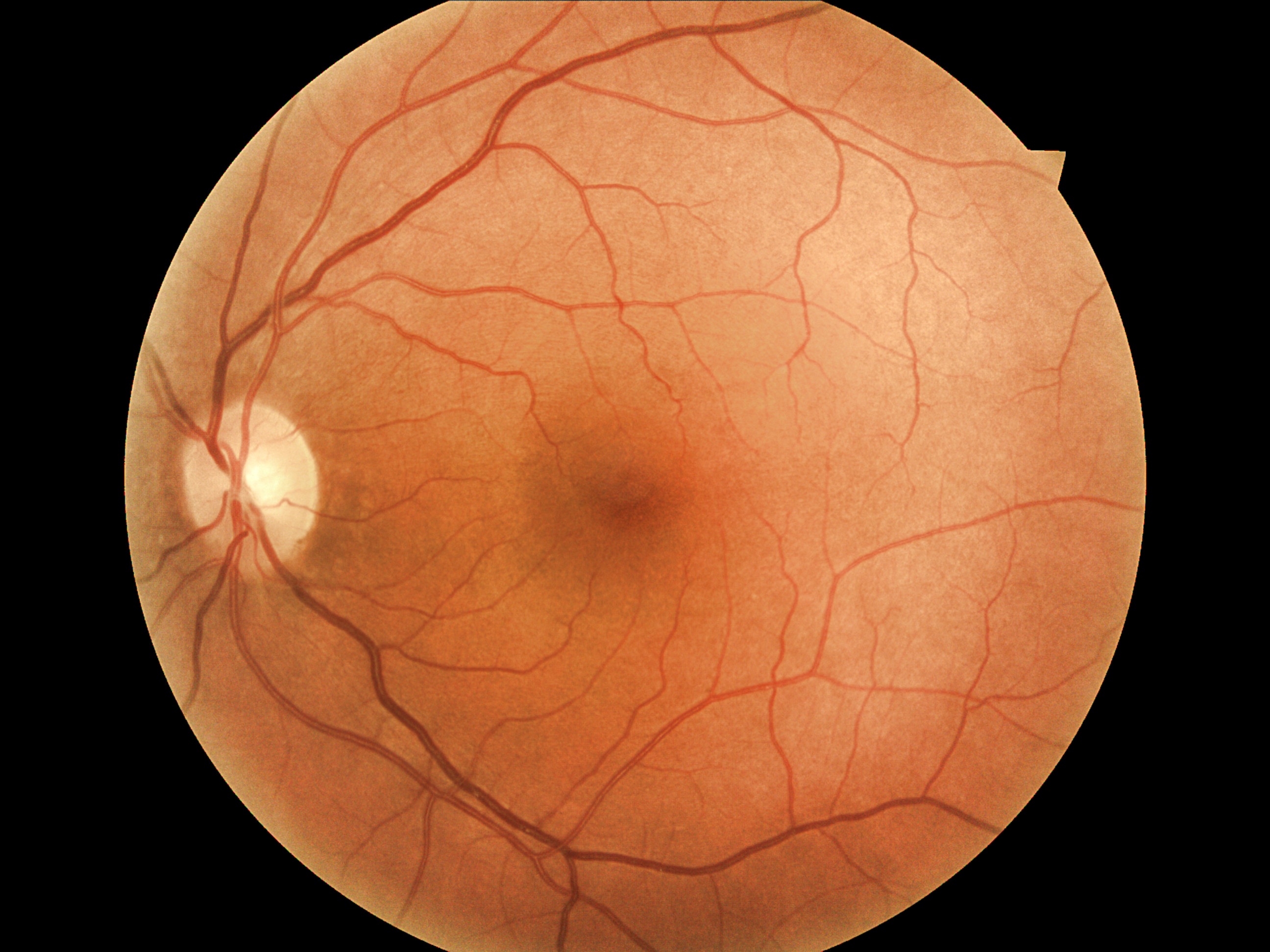}%
}
\subfloat{\includegraphics[width=.15\textwidth]{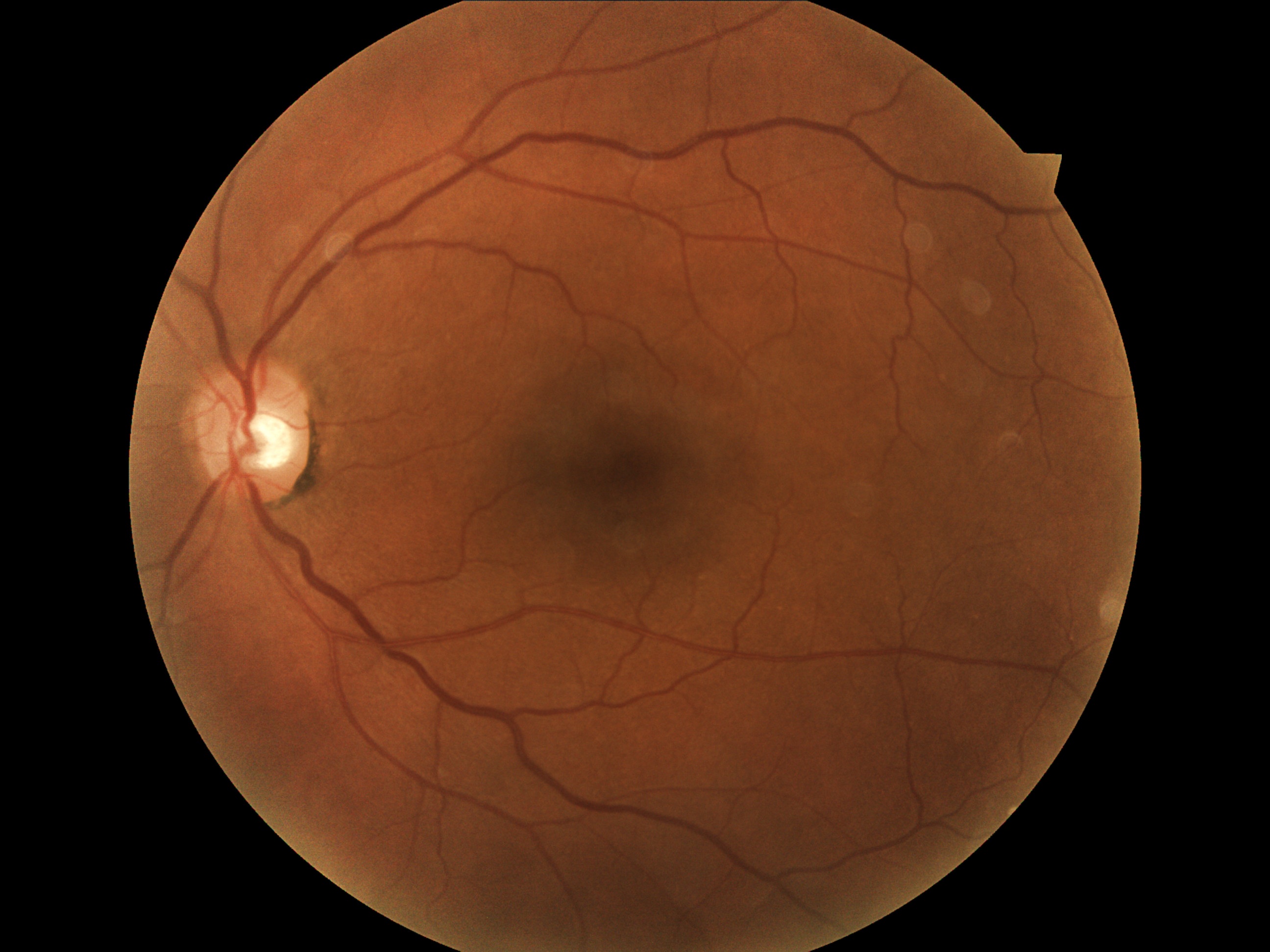}%
}
\caption{Examples of retinal fundus images from the EyePACS dataset. For the left image $(Y,S) = (1,0)$, for the middle $(Y,S) = (0,0)$, and for the right $(Y,S) = (0,1)$.}
\label{dataset}
\end{figure}

We are interested in skin tone for our sensitive variable $S$ with $S = 0$ representing light skin and $S = 1$ dark. However, as skin tone is not included in the dataset, we instead use the Individual Topology Angle (ITA) \cite{wilkes2015fitzpatrick} as a proxy, which was found to correlate with the Melanin Index, frequently used in dermatology to classify human skin on the Fitzpatrick scale.
As in \cite{merler2019diversity,paul2020tara}, we compute ITA via 
\begin{equation}
\mathrm{ITA}=\frac{180}{\pi} \arctan \left(\frac{L-50}{b}\right)
\end{equation}
where $L$ is luminescence and $b$ is ``yellowness'' in CIE-Lab space. We then binarize ITA where an ITA of $\leq 19$ is taken to mean dark skin, as done in \cite{kinyanjui2019estimating,paul2020tara}. Using ITA as a proxy for skin tone has the advantage of being significantly easier to determine compared to the potential issues arising in having a clinician manually annotate images as done in~\cite{burlina2021addressing}. 
Sample images from the EyePACS dataset are shown in Fig.~\ref{dataset}.

\subsection{Metrics}
We use the following metrics to evaluate how well the model performs:

\subsubsection{Measure of Utility}
We use the overall classification accuracy (later denoted $acc$).
\subsubsection{Measures of Fairness} We measure this in multiple ways: a) using the gap in accuracy (denoted $\accgap$) between favored and protected subpopulations; and b) reporting the minimum accuracy across subpopulations (denoted as $\accmin$), which is based on the Rawlsian principle of achieving fairness by maximizing $\accmin$ \cite{rawls2001justice}. Also echoing \cite{skoglund2020}, we measure: c) the adherence to demographic parity via its gap $\dpgap$; and adherence to d) equalized odds via its gap $\eqoddsgap$. The latter two metrics are respectively given by: 
\begin{equation}
\dpgap = \big| P(\hat{Y}=1|S=0) - P(\hat{Y}=1|S=1)\big|
\end{equation}
and
\begin{equation}
\begin{split}
\eqoddsgap = \max_{y \in \{0,1\}} | P(\hat{Y}=1|S=0,Y=y) \\
-P(\hat{Y}=1|S=1,Y=y) |. 
\end{split}
\end{equation}

\subsubsection{Joint Utility-Fairness Measure}  Echoing and comparing with the work in \cite{paul2020tara}, we use a single metric that jointly captures utility and fairness, the Conjunctive Accuracy Improvement ($CAI_{\lambda}$) measure: 
\begin{equation}
CAI_{\lambda} = \lambda (\accgap^b - \accgap^d) + (1 - \lambda) (acc^d - acc^b)
\end{equation} 
where $ 0 \leq \lambda \leq 1 $, and $acc^b$ and $acc^d$ are the accuracy for baseline and debiased algorithms, respectively, while $\accgap^b$ and $\accgap^d$ are gap in accuracy for the baseline and debiased algorithms. In practice, one uses either $\lambda=0.5$ for equal balance between utility and fairness or $\lambda=0.75$ to emphasize fairness.

\subsection{Additional Implementation Details}

We use an isotropic Gaussian distribution for the encoder with mean and variance learned by a neural network, $P_{Z|X} = \mathcal{N}(Z|\mu_\text{enc}(X),\sigma^2_\text{enc}(X)I_d)$, where the mean and variance are two outputs of the encoder, $d$ is the dimension of $Z$, and $I_d$ is the $d$-dimensional identity matrix. The representation is then computed as $Z = \mu_\text{enc}(X) + \sigma_\text{enc}\mathcal{N}(\underbar{0},I_d)$, where $\underbar{0}$ is the all-zero vector of size $d$.

We model the approximation of the representation's marginal as a $d$-dimensional spherical Gaussian, $Q_Z = \mathcal{N}(Z|\underbar{0},I_d)$.
As a result, we calculate the R\'{e}nyi divergence in \eqref{final-loss} between the multivariate Gaussians $P_{Z|X}$ and $Q_Z$ using the closed-form expression derived in \cite{gil2013renyi,burbea1984convexity}).

Finally, as we only use binary values for $Y$, we model $Q_{Y|Z}$ with Bernoulli distributions, $Q_{Y|Z} =$ Bernoulli($Y;f(Z)$) and $Q_{Y|Z,S} =$ Bernoulli($Y;g(Z,S)$) where $f$ and $g$ are auxiliary fully connected networks.

We use ResNet50 as the encoder network while $f$ and $g$ consist of two linear layers followed by a Sigmoid layer. After creating the representation $Z$, we use a logistic regression classifier with default settings 
to predict $Y$ from $Z$. We evaluate accuracy and fairness on these predictions.

\subsection{Results}
We predict $Y=$ DR Status while using $S=$ ITA as the sensitive attribute. We consider the case of an extreme data imbalance where training data is completely missing for one protected subgroup (dark skin individuals) and for a specific value of $Y=1$ (DR-referable individuals). We create a training partition containing both images referable and non-referable for DR of light skin individuals but only non-referable images of dark skin individuals. The goal is for predictions on the missing subgroup to be just as accurate as on the group with adequate training data. This is a problem of both fairness and also \textit{domain adaptation}, and matches an important real world problem where data for dark skin individuals is lacking compared to light skin individuals. 

 For a fair assessment of our method's performance we evaluate on a balanced test set with an equal number of positive and negative examples for both dark and light skin individuals. We use the same partition as in \cite{paul2020tara} to compare with their method. For their method, we report their original CAI scores calculated with respect to their baseline whereas we calculate our CAI scores with respect to results from our own baseline, a ResNet50 network. 

\textit{Hyper-parameter tuning:} We use various combinations of hyperparameters $\beta_1$ and $\beta_2$ varied linearly from 1 to 50 and $\alpha$ varied linearly from 0 to 1, where $\alpha=1$ signifies KL divergence instead of R\'enyi divergence.
As values of $\alpha = 1$ and $\beta_2 = 0$ correspond to the IB method, to compare it with our method we first find an intermediate value of $\beta_1$ that performs well for the IB method, then fix our value of $\beta_1$ to the same value and vary $\alpha$ and $\beta_2$. We compare our method to the CFB in the same way, fixing a value of $\beta_2$ that we use for both the CFB and our method, and then varying $\alpha$ and $\beta_1$. We implement these methods ourselves and also compare to two methods with results taken from \cite{paul2020tara}: adversarial independence (AD) that minimizes conditional dependence of predictions on sensitive attributes with an adversarial two player game and intelligent augmentation (IA) that generates synthetic data for underrepresented populations and performs data augmentation to train a less biased model.

As seen in Table~\ref{tab:eyepacs}, our method mostly outperforms all other methods, showing  improvements in  accuracy  and fairness across nearly all metrics. Usual caution should be exercised in interpretations since -- despite our aligning with data partitioning in~\cite{paul2020tara} -- other variations may exist with~\cite{paul2020tara,deepvariational,skoglund2020} due to non-determinism, parameter setting or other factors. 
%


\begin{table*}[htbp]
\caption{Results for Debiasing Methods on EyePACs}
\scriptsize
\centering
\begin{threeparttable}
\begin{tabular}{l||c|c|c|c|c|c|c}
    \toprule
\multicolumn{1}{l||}{Methods}                               & \multicolumn{1}{l|}{$acc \uparrow$} & \multicolumn{1}{l|}{$\accgap \downarrow $} & \multicolumn{1}{l|}{\begin{tabular}{@{}l@{}} $\accmin \uparrow$ \\  (subpop.) \end{tabular}} & \multicolumn{1}{l|}{$CAI_{0.5} \uparrow$} & \multicolumn{1}{l|}{$CAI_{0.75} \uparrow$} & \multicolumn{1}{l|}{$\dpgap \downarrow$} & \multicolumn{1}{l}{$\eqoddsgap \downarrow$} \\ \hline \hline
Baseline (from ~\cite{paul2020tara}) & 70.0                                & 3.5                                          & 68.3                                                   & -                                         & -                                          & NA                                         & NA                                            \\ \hline
AD ($\beta = 0.5$) (~\cite{paul2020tara})                                        & 76.12                               & 2.41                                         & 74.92 (L)                                          & 3.61                                      & 2.35                                       & NA                                         & NA                                            \\ \hline
IA (~\cite{paul2020tara})                                                        & 71.5                                & 1.5                                          & 70.16 (D)                                           & 1.75                                      & 1.875                                      & NA                                         & NA                                            \\ \hline
Baseline (ours)                                                   & 73.37                               & 8.08                                         & 69.33 (D)                                           & -                                         & -                                          & 28.25                                      & 36.33                                         \\ \hline
IB ($\beta_1$=30) (~\cite{deepvariational})                                          & 74.12                               & 2.08                                         & 73.08 (D)                                           & 3.37                                      & 4.69                                       & 18.58                                      & 20.67                                         \\ \hline
CFB ($\beta_2$=30) (~\cite{skoglund2020})                                        & 77.83                               & 1.66                                         & 77.0 (L)                                           & 5.84                                      & 5.93                                       & \textbf{10.83   }                                   & \textbf{12.5    }                                      \\ \hline
\begin{tabular}{@{}l@{}} RFIB (ours)\\  ($\alpha = 0.8, \beta_1 = 36, \beta_2$ = 30) \end{tabular}          & \textbf{79.42   }                            & 0.5                             & \textbf{79.17 (L)  }                                        & \textbf{6.81   }                                 & \textbf{7.19     }                         & 16.17                                   & 16.67                                          \\ \hline
 
\begin{tabular}{@{}l@{}} RFIB (ours)\\  ($\alpha = 0.8, \beta_1 = 30, \beta_2$ = 50) \end{tabular}                & 79.17                  & 0.83                                         & 78.75 (L)                                 & 6.54                             & 6.91                                      & 17.0                              & 17.83                               \\ \hline
\begin{tabular}{@{}l@{}} RFIB (ours)\\  ($\alpha = 1.8, \beta_1 = 30, \beta_2$ = 17) \end{tabular}                & 78.35                  & \textbf{0.25}                                         & 78.25 (L)                                 & 6.41                             & 7.12                                      & 15.58                             & 15.83                               \\ \hline
\bottomrule
\end{tabular}
\begin{tablenotes}
\item[] {We predict $Y$= DR Status, trained on partitioning with respect to $S=$ ITA, and evaluated on a test set balanced across DR status and ITA. For metrics with an $\uparrow$ higher is better whereas for $\downarrow$ lower is better.  Subpopulation is the one that corresponds to the minimum accuracy, with (D) indicating dark skin and (L) light skin. Metrics are given as percentages.}
\end{tablenotes}
\end{threeparttable}
\label{tab:eyepacs}

\end{table*}

%% file: discussion.tex
\section{Discussion}
This study proposed RFIB, a new variational approach that encodes trade-offs between different principles of fairness along with utility and compactness. Compared to prior work which incorporates a single definition of fairness, our approach has the potential benefit of allowing ethicists and policy makers to specify softer and more balanced fairness requirements that may lie between multiple hard fairness requirements. Our work opens the way to more future studies expanding on this idea. 

 In additional experiments we evaluated the performance of our method when the fairness model for EyePACS was developed vis-a-vis the ITA protected factor but  was then tested on a dataset where the protected factor was race, as annotated by clinicians. The results demonstrate the ability of our method to perform well in this type of protected factor domain adaptation problem, where our approach outperformed all other methods. This is particularly important in a setting where the actual protected factor is not revealed for privacy reasons. Regarding addressing methods that are both fair and private, the type of information blinding approach pursued here may have implications for models that protect against attribute inference attack, another possible future direction.
 
 Overall, evaluation of the method showed benefits vis-a-vis
 methods of record including IB, CFB, and other augmentation or adversarial debiasing techniques, with the potential to further improve results by combining our method with other post-processing, pre-processing and data intervention methods.